\documentclass[10pt,conference]{IEEEtran}

\usepackage[cmex10]{amsmath}
\usepackage{algorithmic}
\usepackage{algorithm}

\usepackage{mdwmath}
\usepackage{mdwtab}
\usepackage{eqparbox}
\usepackage[caption=false,font=footnotesize]{subfig}
\usepackage{fixltx2e}
\usepackage{stfloats}
\usepackage{url}
\usepackage{graphics}

\input epsf
\usepackage{multirow}
\usepackage{graphicx}
\usepackage{authblk}
\usepackage{amsmath}
\usepackage{amssymb}
\usepackage{setspace}	
\usepackage{array}
\usepackage{enumitem}
\usepackage{tabu}
\usepackage{cite}
\usepackage{multibib}
\newcommand{\eat}[1]{}

\begin{document}

\title{Application of Hierarchical Temporal Memory Theory  for Document Categorization}

\author[1]{Deven Shah}
\author[2]{Pinak Ghate}
\author[3]{Manali Paranjape}
\author[4]{Amit Kumar}

\affil[1]{College of Engineering, Pune, India, \textit{shahdeven04@gmail.com}}
       
\affil[2]{College of Engineering, Pune, India, \textit{pinakghate@gmail.com}}

\affil[3]{College of Engineering, Pune, India, \textit{paranjape.manali@gmail.com}}

\affil[4]{Maker's Lab, Tech Mahindra Ltd., India, \textit{ak00494790@techmahindra.com}}
       
\maketitle

\begin{abstract}
The current work intends to study the performance of the Hierarchical Temporal Memory(HTM) theory for automated classification of text as well as documents. HTM is a biologically 
inspired theory based on the working principles of the human neocortex. 
The current study intends to provide an 
alternative framework for document categorization using the 
Spatial Pooler learning algorithm in the HTM Theory. As HTM accepts only a stream of binary data as input, Latent Semantic Indexing(LSI) technique is used for 
extracting the top features from the input and converting them into binary 
format. The Spatial Pooler algorithm converts the binary input into sparse 
patterns with similar input text having overlapping spatial patterns making it 
easy for classifying the patterns into categories. The results obtained prove that HTM 
theory, although is in its nascent stages, performs at par with most of the popular machine learning based classifiers.
\end{abstract}
	
\IEEEoverridecommandlockouts

\smallskip
\begin{IEEEkeywords}
Hierarchical Temporal Memory, Document Categorization, Machine Learning, Spatial 
Pooler, Latent Semantic Indexing, NuPIC, Supervised Learning
\end{IEEEkeywords}

\IEEEpeerreviewmaketitle

\section{Introduction}
One of the elemental forms of document processing includes classification. 
Since the last couple of years, it is in demand because of the increasing 
availability of data in digital format which has resulted into the  requirement of systematization of  
that data. Manual organization of huge data can be tedious if strict time 
constraints are set, increasing the necessity of automated document classification. The contexts of words in the documents play a very important role in deciding the category of the document. The human brain is very effective in consideration of contexts in the incoming information for taking the appropriate action. 

\smallskip

The principles of HTM theory can be used to meet the requirements of organizing of data.  
HTM 
takes inspiration from the mammalian brain which has been evolving over 
millions of years and is able to process data efficiently. As HTM is 
biologically plausible, it is based on simple rules and not complex mathematics. HTM theory is being developed by a US based company called Numenta, Inc.
\smallskip

\section{Related Work}
Some of the conventional methods for text/document classification are mentioned below:
\subsection{Naive Bayes}
The Naive Bayes classifier is a probabilistic classifier and is based on the Bayes theorem. It works well with small samples of data. The posterior probability of a particular document belonging to various classes is calculated. The document is assigned to the class with the highest posterior probability. The Naive Bayes classifier assumes strong independence between the features. This is a major limitation of this classifier and hence has low performance in cases where the features are correlated \cite{comparison}.

\subsection{Support Vector Machines}
Support Vector Machines (SVMs) are supervised machine
learning algorithms. In case of a multi class problem, first
the problem has to be decomposed into two separate class
problems as SVM can work only with binary classification
problem. They will probably give poor results when total
number of samples are very less than the total number of
features. In comparison with decision making classifier and
logistic regression, SVM takes more time for computation \cite{comparison}.

\subsection{K-Nearest Neighbour}
K-Nearest Neighbour (KNN) is used for classification of objects by calculating the distance of training samples from each object. KNN classification is a simple and widely used approach for text classification. However, it is computationally intensive and
classification time is high \cite{comparison}. Also, it is difficult to find the ideal value of k \cite{knn}.

\subsection{Convolutional Neural Network}
Convolutional Neural Network (CNN) works well with static text classifications. CNN is a type of feed forward neural network, comprising of neurons with trainable weights and biases. CNN comprises of a number of convolutional layers with nonlinear activation functions like ReLU or tanh applied to the results. CNN suffers from the limitations of the requirement of large data and big processing power to be able to predict accurately \cite{cnn}.

HTM theory is primarily used for Classification, Prediction and Anomaly Detection purposes. One of its application for Classification is mentioned below:

\subsection{Land - forms classification}
As HTM based models have a common learning algorithm, it can be used for classifying images. HTM theory has been used for classifying different land-forms like trees, roads, buildings and farms using the images obtained from satellites. The framework used achieved an accuracy of 90.4\%,\cite{perea2009application}  which is at par with the conventional machine learning techniques for image classification.

Since HTM theory can be used for image classification purposes, it can hold a promise to classify text/documents.

\section{Overview of Hierarchical Temporal Memory}
HTM is a theory which seeks to apply the 
structural as well as algorithmic properties of the neocortex to machine 
learning problems \cite{hawkins2010hierarchical}. The neocortex proves to be the center of intelligence in the 
mammalian brain. It is responsible for processing complex activities such as 
communication, planning and prediction. Structurally, neocortex is a 2 mm thick 
tissue divided into a number of different regions. A region is a network of 
interconnected neurons \cite{hawkins2010hierarchical}. This attributes to the presence of input connections 
from different sensory organs \cite{hawkins2006hierarchical,onintelligence} like eyes, ears etc. The term “Hierarchical” in the theory is owing to 
the fact that, HTM network contains a hierarchy of levels arranged in the 
pyramid-like structure. These levels are present in the form of regions that 
are again composed of columns which finally consist of neurons. These neurons need 
not be physically arranged in a hierarchy, but are logically arranged in the 
hierarchical format. The lower levels in hierarchy represent data having lower 
abstraction/complexity. As we go higher in the hierarchy, the data abstraction stored in the memory increases. Time plays a crucial role in the 
way data is stored in mammalian brain. “Temporal” implies that the HTM network takes 
into consideration the sequence of the incoming data. A continuous stream of 
input data is aptly learned as spatial and temporal sequences.
\smallskip

A remarkable property of the neocortex is that the input from all the sensory 
organs is processed in the same manner. Hence, it has a common learning algorithm 
for inputs from all types of sensory organs \cite{hawkins2010hierarchical}. 

\begin{figure*} [htb!] 
\includegraphics[width=1\textwidth]{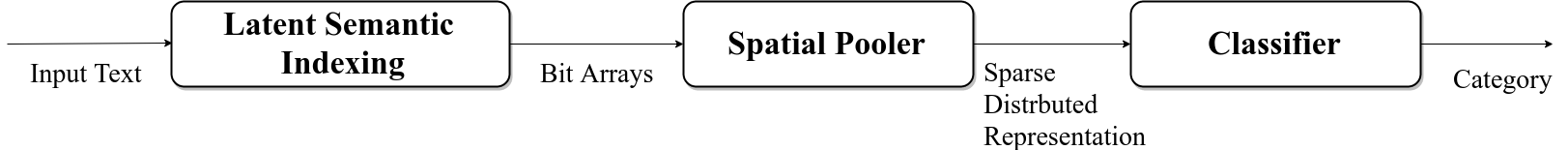}
\caption{\sl System Architecture Diagram
\label{fig:implementation}}
\end{figure*}

\subsection{Structure of a Neuron}

Inside the mammalian brain, neurons play a central role in information handling. Some 
relevant parts of the neuron for our study are mentioned below. 

\subsubsection{Proximal Dendrites}
Proximal dendrites are in close proximity to the cell body. The proximal 
dendrites are connected directly to the inputs from the sensory 
organs.

\subsubsection{Distal Dendrites}
Distal dendrites are the ones that are afar from the cell body. The distal 
dendrites have connections with various other neurons in the neocortex. Majority of the 
connections to the axon are from distal dendrites as compared to the 
connections made by proximal dendrites \cite{synapse}.

\subsubsection{Synapse}
A synapse is a connection between an axon of one neuron and dendrite of the 
other. The ongoing process of breaking and reforming these synapses between 
cells results in learning of new data and thus gradually forgetting the old 
one.\smallskip

There is a permanence value associated with every synapse and a threshold 
linked with every neuron. Thus, for a neuron to get activated, the total number 
of synapses with permanence values higher than the threshold value must be more 
than the stimulus threshold.

\subsection{Sparse Distributed Representation}
Though the neocortex contains billions of neurons in highly interconnected 
manner, only a tiny fraction of them are active for a particular input \cite{sdr1}. Hence, only small percentage of active neurons are responsible for 
representing the input information. This is called as Sparse Distributed Representation (SDR). 
Even though single activated neuron has the potential to convey some meaning, 
the full information can only be conveyed when it is interpreted within the 
context of other neurons.
As the information is spread across a tiny percentage of the active bits, SDRs are more noise tolerant than dense representations, making them ideal for text processing.

\subsection{Spatial Pooler}
HTM includes two important parts - Spatial Pooler (SP) and Temporal Pooler (TP). Spatial Pooler, also known as Pattern Memory, has been emphasized in this study.
\smallskip

The neurons in the neocortex are arranged in columns, which represent features of the input. Every neuron in a particular column, which represent different context for an input, is 
connected to specified number of bits in the input bit array. The selection of 
bits to be connected to the neurons in a particular column is random. The bits 
which are connected to a particular column are known as a potential pool of 
that particular column. Connections between input bit and the column neuron is 
called as a synapse. Every synapse has a value associated with it known as 
permanence value similar to that of a mammalian brain. Permanence value is always in the range of 0 and 1. There is a threshold value associated with synapse's permanence.

If the permanence value of a synapse associated to an input bit is greater than the threshold, the activation of the column of neurons 
is influenced by the input bit. The permanence value of a synapse is adjusted in the learning phase.

The main role of SP in HTM is finding spatial patterns in the input data. It is decomposed into three stages:
\medskip
\subsubsection{Overlap}
In this stage, overlap score of each column is calculated. Overlap score is 
 the count of active bits in the potential pool of a particular 
column having permanence value greater than the threshold. 
\smallskip
\subsubsection{Inhibition}
The columns are sorted according to their overlap scores from highest to lowest. A particular fraction (in our study, 0.5\%, Table \ref{spatial_pooler_paramters}, $NumActiveColumnsPerInhArea$) of the top columns is selected (also called as active columns or the winning columns) for the learning phase. Rest other columns are inhibited from learning.
\smallskip
\subsubsection{Learning}
During Learning, the permanence value of the synapses in the potential pool of the winning 
columns is incremented (by synPermActiveInc, Table \ref{spatial_pooler_paramters}) or decremented (by synPermInactiveDec, Table \ref{spatial_pooler_paramters}). When the active column is connected to an 
active bit then the permanence value of the synapse corresponding to that 
active bit is incremented. However, when the active column is connected to an
inactive bit then the permanence value of the synapse corresponding to that 
inactive bit is decremented. This is the result of column expecting that bit to be active. The synapse permanence is decremented as a 
punishment. 

\section{Implementation}

The flowchart in figure \ref{fig:implementation} is our high-level 
architecture diagram for document categorization. As the mammalian brain requires electrical 
signals for learning, the learning algorithm i.e., Spatial Pooler also requires 
bit patterns for processing. So, to convert text into bit arrays, Latent Semantic 
Indexing (LSI) technique is used, which converts semantically similar sentences 
into similar bit arrays. These bit arrays (which need not be sparse)
are fed to the Spatial Pooler where it simulates the working of neurons in 
the brain and gives SDR as the output. The 
active bits in the SDR represent the neurons 
which get activated in the Spatial Pooler. Since semantically similar 
text belong to the same category, it is easy to classify the 
text into different categories.
\medskip
\subsection{Latent Semantic Indexing}
As HTM theory is modelled after the mammalian brain, its input also should be in 
accordance with the input format received by the brain. The brain receives 
input in the form of electrical signals which correspond to bit arrays. Latent Semantic Indexing(LSI) helps in determining hidden 
features in documents \cite{papadimitriou1998latent}. Thus the technique is used to extract the 
contextual-usage meaning of words from the documents\cite{lsa3}. 
The LSI framework consists of 3 steps which are mentioned below.
\smallskip

\subsubsection{Preprocessing of input data}
In the initial step, the input text is tokenized and stopwords are removed from 
every document of the corpus\footnote{We have used the wikipedia language corpus as it 
includes a large vocabulary which is useful for generic datasets. The corpus 
will change if the dataset is in a language other than English or contains a 
large number of words which are not present in the vocabulary.}. Each term in 
the text is then represented as a tuple containing term-id and term frequency. 
A matrix is created in which the rows denote the unique terms and the 
columns denote the documents. Every cell denotes the term count in the corresponding document.
The matrix of term-frequency counts obtained from the term document matrix is 
then modified using the TF-IDF technique so as to give more weight to rare 
terms compared to common terms across documents and also to frequently occurring 
terms in a particular document.   
The formula for weighing each term can be represented as,

\begin{figure}
\begin{centering}
\includegraphics[width=1\linewidth]{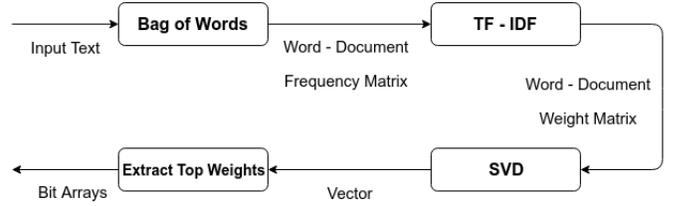}
\caption{\sl Latent Semantic Indexing framework
\label{fig:lsa_framework}}
\end{centering}
\end{figure}

\begin{gather}
  Document Term Weight = f_{t,d} \times \ln(N / n_t) \\
\intertext{Where:}
  \begin{tabular}{>{$}r<{$}@{\ :\ }l}
    f_{t,d} & count of term~$t$ in document~$d$ \\       
    N & the total count of documents \\
    n_t & the count of documents having term~$t$ \\
  \end{tabular}\nonumber
\end{gather}

\smallskip

The term-document matrix gets modified to contain weights of each term 
in a given document. The dimensionality reduction of this matrix is done using 
Singular Value Decomposition (SVD).
\smallskip
\subsubsection{Singular Value Decomposition}
LSA uses SVD for generating the vectors of a particular text \cite{lsa1, lsa2}. The matrix $X$(term-document) is used to calculate two matrices. These are,

\begin{gather}
  Y = X^TX \\
  Z = XX^T \\
\intertext{Where:}
  \begin{tabular}{>{$}r<{$}@{\ :\ }l}
    $X$ & term - document matrix \\
	$Y$ & document - document matrix \\       
    $Z$ & term - term matrix \\
  \end{tabular}\nonumber
\end{gather}

After finding eigenvectors of $Y$ and $Z$ matrices, we get left singular matrix, $L$ 
and right singular matrix, $R$ respectively. Thus, term - document matrix, $X$, 
 is divided into unique combination of three matrices as follows -

\eat{
\begin{gather}
 X = L \Sigma R^T \\
\intertext{Where:}
  \begin{tabular}{>{$}r<{$}@{\ :\ }l} 
  \multirow{2}{*}{$L$} & is the left singular vector matrix representing \\
  & weights of a term for corresponding concepts. \\
  \multirow{2}{*}{$R^T$} & is the transpose of the right singular vector matrix 
\\
  & representing  weights of documents belonging to particular concepts. \\
\smallskip
\multirow{2}{*}{$\Sigma$} & is the diagonal matrix representing \\
  & weights of the concepts found in the text\\
  \end{tabular}\nonumber
\end{gather}
}

\begin{gather}
\label{4}
 X = L \Sigma R^T \\
\intertext{Where:}
  \begin{tabular}{>{$}r<{$}@{\ :\ }l} 
  $L$ & Term - Concept weight matrix \\
  R^T & Concept - Document weight matrix \\
  \Sigma & Diagonal matrix representing concept weights \\
  \end{tabular}\nonumber
\end{gather}

$\Sigma$ is calculated by taking the square root of the eigenvalues of matrix $Y$. 
\smallskip

To reduce the dimensionality of the matrices in equation \ref{4}, top $k$ concepts are selected and thus 
matrix $X$ is approximated as,

\begin{gather}
 X_k = L_k \Sigma_k R_k^T
\end{gather}

In our study, $k$ is taken to be 400 in order to consider top 400 concepts. This 
marks the end of the training phase. 

In the testing phase, after generating 
weight matrix using the Term Frequency - Inverse Document Frequency (TF - IDF) 
model, input text gets converted into a query matrix, $Q$. This matrix $Q$ is then 
multiplied with matrices $L_K$ and $\Sigma_k$ to generate new query vectors calculated as follows:

\begin{gather}
  New Query Vectors = Q L_k \Sigma_k
\end{gather}

\subsubsection{Extraction of top features}

The query vectors are converted into bit arrays of size 400. 
The indices of the top 40 features from the query vectors represent the '1's in the bit arrays and the indices of the remaining features represent '0's.

\subsection{Spatial Pooler}
The bit arrays from the LSI encoder are then passed to the Spatial Pooler for 
learning. The Spatial Pooler gives similar Sparse Distributed Representations 
(SDRs) for similar input text. The major parameters of the Spatial Pooler which 
significantly affect the accuracy of our model are mentioned in Table \ref{spatial_pooler_paramters}.

\medskip

\begin{table}[htbp]
\caption{Spatial Pooler Parameters}
\label{spatial_pooler_paramters}
\begin{center}

\begin{tabular}{|l|c|}
	\hline
	\textbf{Parameters} & \textbf{Values}\\
	\hline
    inputDimensions & 400\\
	\hline
    columnDimensions & 20000\\
	\hline
    potentialRadius & 200\\
	\hline
    numActiveColumnsPerInhArea & 100\\
	\hline
    synPermActiveInc & 0.01\\
	\hline
    synPermInactiveDec & 0.008\\
	\hline
\end{tabular}
\end{center}
\end{table}

\smallskip
The active indices of the SDR are then fed to the Classifier.

\subsection{Classifier}
\begin{figure}  
\includegraphics[width=0.9\linewidth]{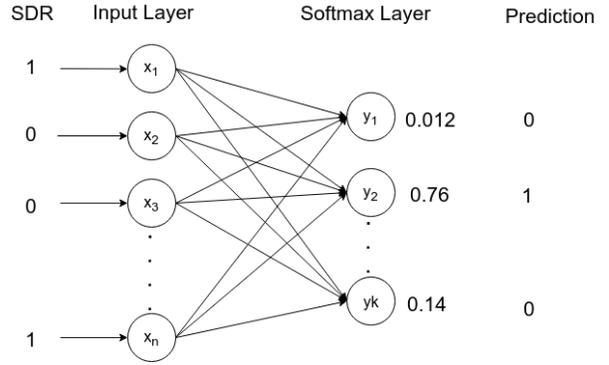}
\caption{\sl Classifier
\label{fig:classifier}}
\end{figure}

In order to predict target class labels, a sequence of N-dimensional SDRs is 
assigned to a set of k class labels. The Classifier makes 
use of a single layer feed forward neural network. In the figure 
\ref{fig:classifier}, the number of output neurons is equal to the number of 
predefined categories. The number of input neurons is equal to the number of 
bits in any SDR.

The algorithmic description of the classifier is as follows,
\smallskip
\subsubsection{Matrix Initialisation}
Since all classes have an equal chance of occurrence before learning, all 
values in the weight matrix are initialised to zero. 
\smallskip
\subsubsection{Inference}
In this phase, the predicted class probabilities for each input pattern are 
calculated. The calculations include two steps as mentioned below.
\smallskip
\setlength{\parindent}{4ex}

\textit{i) Weighted sum :} Weighted sum of the input is calculated for each output neuron to determine the activation levels of the neuron. 
Activation level of an output neuron can be determined by the summation of the product of all the input bits to the input layer neurons with the weights of its corresponding connections to the output layer neuron. 
The formula for the activation level being,

\begin{gather}
 a_j = \sum_{i=1}^{N} w_{ij} \times x_i \\
\intertext{Where:}
  \begin{tabular}{>{$}r<{$}@{\ :\ }l} 
  a_j & activation level of the $j^{th}$ output layer neuron \\
  n & number of input layer neurons\\
  \eat{w_{ij} & weight of the connection from the $i^{th}$ input layer  
\\ & neuron to the $j^{th}$ output layer neuron \\}  
   \multirow{2}{*}{$w_{ij}$} & weight of the connection from the $i^{th}$ input 
neuron \\
  & to the $j^{th}$ output neuron. \\
  x_i & Input bit value. It is either 0 or 1. \\
  \end{tabular}\nonumber
\end{gather}

\par

\setlength{\parindent}{4ex}

\textit{ii) Softmaxing the activation levels :} The probability distribution of 
the categories is calculated by exponentiating and normalizing the activation 
levels of the neurons in the output array using the softmax function. The 
formula for the probability distribution being,

\begin{gather}
 P\left[ C_k | x, w\right] = y_k = \frac{e^{a_k}}{\sum_{i=1}^{k} e^{a_i}} \\
\intertext{Where:}
  \begin{tabular}{>{$}r<{$}@{\ :\ }l} 
  y_k & Probability of predicting the category index $k$. \\
  k & Number of predefined categories. \\
  \end{tabular}\nonumber
\end{gather}

\par

\subsubsection{Learning}
During each iteration, the classifier makes a prediction of the category index 
of a given SDR. This prediction is of the form of the probability distribution 
over different category indexes.
The connection weights are updated to learn and improve the prediction results. Connection weights are adjusted only for the active bits. 
The Connection Weights are determined using maximum likelihood estimation (MLE) on independent 
input SDRs. Since, the SDRs are independent of each other, they would satisfy 
the following equation.

\begin{gather}
 P\left[ z^1, z^2, ..., z^t\right] =  \prod_{t} P\left( z^t | x^t, w\right) \\
 x^t = (x^{t}_1, x^{t}_2, x^{t}_3, ..., x^{t}_N) \\
\intertext{Where:}
  \begin{tabular}{>{$}r<{$}@{\ :\ }l} 
  z^t & actual category index of $t^{th}$ SDR. \\
  x^t & Sparse Distributed Representation. \\
  x^{t}_1 & first bit of the $t^{th}$ SDR. \\
  w & Connection weights. \\
  \end{tabular}\nonumber
\end{gather}

A value of $w$ is selected so that likelihood gets maximized. The loss function to select $w$ \cite{neuneier2012train}, is as follows : 

\begin{gather}
 L = -\ln \left(  \prod_{t} p\left(y_t | x_t, w\right)\right) \\
= -\Sigma_{t}\ln P\left(y_t | x_t, w\right)
\end{gather}

Gradient descent is used is to minimize the loss function.

\begin{gather}
\frac{\partial L}{\partial w_{ij}} = \frac{\partial L}{\partial a_j} \times 
\frac{\partial a_j}{\partial w_{ij}} \\
=\left(y_j - z_j\right)x_i \\
\intertext{Where:}
  \begin{tabular}{>{$}r<{$}@{\ :\ }l} 
  y_j & Predicted probability of $j^{th}$ category index. \\
  z_j & Actual probability of $j^{th}$ category index. \\
  x_i & Input bit value to the $i^{th}$ input neuron. \\
  \end{tabular}\nonumber
\end{gather}

Error in connection weight between $i^{th}$ input neuron to the $j^{th}$ output neuron is,

\begin{gather}
Error_{ij}\ for\ active\ input\ bits = y_j - z_j
\end{gather}

\begin{gather}
update_{ij} = \alpha \times error_{ij} \\
\intertext{Where:}
  \begin{tabular}{>{$}r<{$}@{\ :\ }l} 
  \alpha & Learning rate. \\
  \end{tabular}\nonumber
\end{gather}

The value $update_{ij}$ is used to update the connection weight between the 
$i^{th}$ input neuron to the $j^{th}$ output neuron using the formula,

\begin{gather}
w_{new_{ij}} = w_{old_{ij}} + update_{ij}
\eat{\intertext{Where:}
  \begin{tabular}{>{$}r<{$}@{\ :\ }l} 
  \alpha & Learning rate. \\
  \end{tabular}\nonumber}
\end{gather}

But, if we just want to update connection weights for the bits which are active 
we multiply $update_{ij}$ with $x_i$.

\begin{gather}
w_{new_{ij}} = w_{old_{ij}} + update_{ij} \times x_i
\intertext{Where:}
  \begin{tabular}{>{$}r<{$}@{\ :\ }l} 
  w_{new_{ij}} & updated weight of the connection from $i^{th}$ input \\
  &  neuron to the $j^{th}$ output neuron \\
  \end{tabular}\nonumber
\end{gather}

The output layer neuron with the highest probability represents the category index of 
the input text.

\section{Results}
Many experiments were performed to test the accuracy and performance of 
our model. We selected two standard datasets for document classification,
namely, 20 Newsgroup dataset from the sklearn dataset repository and Movie 
Reviews dataset from the NLTK corpus repository. The datasets were 
split into train set and test set in the ratio 9:1. The classification framework used in this study gives comparable accuracies with the models mentioned in the table \ref{tp_rate} on the same datasets. 
\smallskip

\begin{table}[htbp]
\caption{TRUE POSITIVE RATE}
\label{tp_rate}
\begin{center}

\begin{tabular}{ | m{6.5em} | m{2cm}| m{2cm} | } 

\hline
\bfseries Classification Techniques & \bfseries 20 newsgroup & \bfseries Movie 
Reviews \\ 
\hline
SVM\cite{svmmr} & ---- & 84.40\% \\ 
\hline
Decision Trees\cite{decisiontreesmr} & ---- & 61.10\% \\ 
\hline
Naive Bayes\cite{naivebayes20,naivebayesmr} & 86.00\% & 62.35\% \\ 
\hline
B-Tree\cite{btree20} & 82.64\% & ---- \\ 
\hline
Bayesian Networks \cite{bayes_net}& 78.58\% & ---- \\
\hline
HTM & 83.19\% & 73.60\% \\
\hline

\end{tabular}
\end{center}
\end{table}

\smallskip

\section{Conclusion and Future Scope}

This paper puts forward the results of using the Hierarchical Temporary Memory 
model for document categorization. The results prove that the HTM model gives 
an accuracy comparable to the conventional techniques used for text 
classification. 
The number of columns and the SDR sparsity has a significant effect on the 
performance of the spatial pooler. As per our model, The optimal values of the 
number of columns was 20,000 and the sparsity was 0.5\%.

The main advantages of this model are: a limited number of parameters, can be 
trained on small\cite{synapse} corpus and faster training.

In future, we plan to modify the encoding process of our model and also 
incorporate the Temporal Pooler which can help to increase the accuracy of the 
model.

\section{Acknowledgement}
We are grateful to Mr. Nikhil Malhotra of Maker's Lab, Tech Mahindra Ltd. and Mr. Satish Kumbhar of College of Engineering, Pune, for guiding us through the research.

\nocite{*}

\bibliographystyle{IEEEtran}
\bibliography{IEEEabrv,ref,refmanali,ref2,ref3,related_work,to_add}

\end{document}